\documentclass{article}

\usepackage[preprint]{neurips_2024}

\usepackage[utf8]{inputenc} 
\usepackage[T1]{fontenc}    
\usepackage{hyperref}       
\usepackage{url}            
\usepackage{booktabs}       
\usepackage{amsfonts}       
\usepackage{nicefrac}       
\usepackage{microtype}      
\usepackage{xcolor}         
\usepackage{colortbl}

\usepackage{kotex}
\usepackage{booktabs} 
\usepackage{subcaption}
\usepackage{wrapfig}
\usepackage{enumitem}
\usepackage[hang, flushmargin]{footmisc}
\usepackage{hyperref}

\usepackage{multirow}
\usepackage{graphicx}
\usepackage[normalem]{ulem} 
\useunder{\uline}{\ul}{} 
\usepackage{adjustbox}
\usepackage{array}
\usepackage{tikz}
\usetikzlibrary{positioning,fit,shapes,arrows.meta}
\usepackage{amsmath}

\usepackage{etoolbox}
\makeatletter
\patchcmd{\@maketitle}
  {\@bottomtitlebar}
  {\@bottomtitlebar\vskip 0.2in}
  {}{}
\makeatother

\title{EXAONE Path 2.5: Pathology Foundation Model \\ with Multi-Omics Alignment}

\author{
\shortstack{
Juseung Yun \quad Sunwoo Yu \quad Sumin Ha \quad Jonghyun Kim \\ Janghyeon Lee \quad Jongseong Jang \quad Soonyoung Lee
} \\ [0.25em]
LG AI Research
}

\begin{document}

\maketitle

\begin{abstract}

Cancer progression arises from interactions across multiple biological layers, especially beyond morphological and across molecular layers that remain invisible to image-only models.
To capture this broader biological landscape, we present EXAONE Path 2.5, a pathology foundation model that jointly models histologic, genomic, epigenetic and transcriptomic modalities, producing an integrated patient representation that reflects tumor biology more comprehensively. 
Our approach incorporates three key components: (1) multimodal SigLIP loss enabling all-pairwise contrastive learning across heterogeneous modalities, (2) a fragment-aware rotary positional encoding (F-RoPE) module that preserves spatial structure and tissue-fragment topology in WSI, and (3) domain-specialized internal foundation models for both WSI and RNA-seq to provide biologically grounded embeddings for robust multimodal alignment.
We evaluate EXAONE Path 2.5 against six leading pathology foundation models across two complementary benchmarks: an internal real-world clinical dataset and the Patho-Bench benchmark covering 80 tasks. Our framework demonstrates high data and parameter efficiency, achieving on-par performance with state-of-the-art foundation models on Patho-Bench while exhibiting the highest adaptability in the internal clinical setting. These results highlight the value of biologically informed multimodal design and underscore the potential of integrated genotype-to-phenotype modeling for next-generation precision oncology.
\end{abstract}

\section{Introduction}

The rapid advancement of computational pathology has unlocked new avenues for understanding cancer biology through large-scale digital image analysis~\citep{remedis, virchow_ref1, HIPT, phikon}. Whole-slide images (WSI) now serve as powerful representations of tumor morphology, capturing spatial phenotypes across the heterogenetic tissue region using self-supervised learning frameworks~\cite{UNI, lunit, pyeon2025exaone, ctranspath, chief, gigapath}. However, morphology alone remains insufficient for explaining disease mechanisms, as it does not directly reveal the genetic, epigenetic, and transcriptional alterations driving tumor progression~\cite{hanahan2011hallmarks}.

To this end, recent studies have begun incorporating spatial transcriptomics or pathology reports to enrich WSI-derived representations~\citep{titan, jaume2024transcriptomics, prism, threads}. Building on this direction, we propose a multimodal framework that jointly integrates five complementary molecular and histopathologic modalities: \textbf{WSI, bulk RNA-seq, single-nucleotide polymorphisms (SNP), copy-number variations (CNV), and DNA methylation}. Each modality captures a different layer of tumor biology, and their unified modeling provides a more complete and biologically grounded patient representation~\citep{heo2021integrative, ing2025integrating}. By constructing a shared latent space spanning genomic, epigenetic, transcriptomic, and phenotypic dimensions, we aim to move toward explainable multimodal oncology.

\begin{figure}[h]
    \centering  \includegraphics[width=0.8\linewidth]{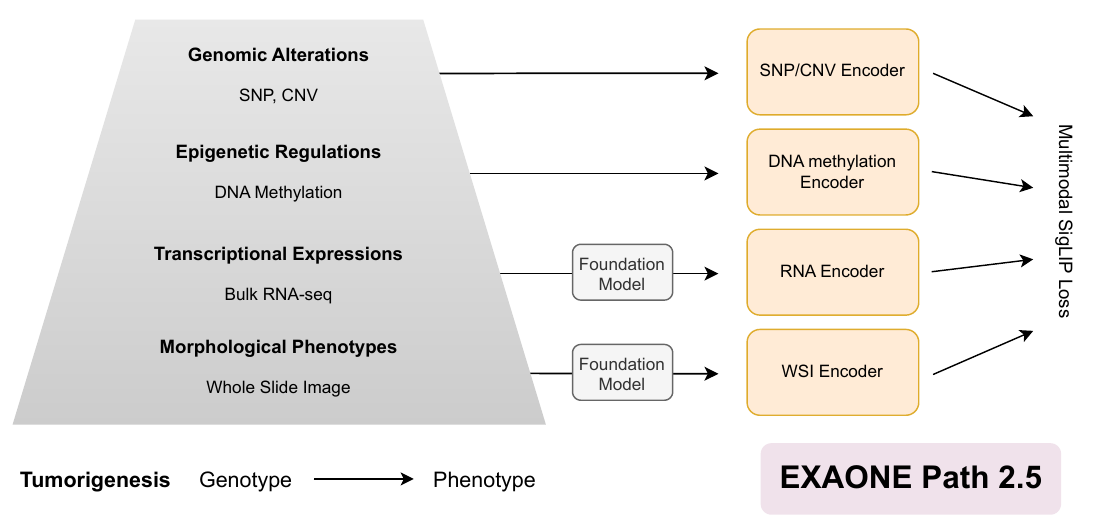}
    \caption{Overall scheme of EXAONE Path 2.5. Whole-slide images are aligned with multi-omics data using a multimodal SigLIP loss that enables all-pairwise cross-modal alignment. By jointly modeling multiple layers of tumor biology from genotype to phenotype, our model produces biologically enriched histopathology representations.}
    \label{fig:modality_exaonepath}
\end{figure}

Although these modalities are correlated, they encode largely orthogonal biological signals. SNPs and CNVs describe the genomic alterations initiating or promoting tumorigenesis~\citep{jain2009role, shields2005factors}. These changes often shape epigenetic regulation through DNA methylation~\citep{kulis2010dna}, which in turn modulates transcriptional programs observed in bulk RNA-seq~\citep{razin1991dna}. WSI then captures the macroscopic phenotype that emerges from these molecular processes. Integrating these five layers of tumorigenesis—from genotype to phenotype—offers a comprehensive view of cancer that cannot be achieved by any single modality alone~\citep{dong2025integration}.

To realize this integration, we design a \textit{pathology-oriented multimodal alignment framework} tailored to real-world clinical variability and missing-modality patterns.

\begin{itemize}
    \item Multimodal SigLIP Loss: We propose a modified SigLIP contrastive loss \cite{siglip} to support efficient \textit{all-pairwise contrastive learning} across the five modalities. All modality pairs are projected using its respective adapter, allowing them to capture different semantic aspects of importance to another modality. This objective encourages each modality to encode complementary biological information while preventing collapse toward any single, easier modality. 
    
    \item Fragment-Aware Rotary Position Encoding: Instead of treating WSI features as unordered tile sets, we preserve spatial structure by \textit{rotary positional embeddings with fragment-aware attention masking}. This allows the model to differentiate distinct tissue fragments on a slide and capture region-level histologic patterns (e.g., variations in tumor grade, stromal composition, or immune infiltration) that often reflect underlying molecular states.

    \item Internal Foundation Models for WSI and RNA-seq: We leverage \textit{internal foundation models} that were pretrained on histopathology and transcriptomic datasets independently. These domain-specialized foundation models provide stable, biologically meaningful embeddings that improve both the robustness and effectiveness of multimodal contrastive alignment.
\end{itemize}

This domain informed design yields strong resource efficiency, enabling the model to match or surpass state-of-the-art performance across diverse tasks with substantially fewer training samples and parameters.

\section{Methods}

\subsection{Multimodal SigLIP Loss} \label{sec:siglip}
Standard CLIP-based losses assume that there is only one correct positive per row. As the softmax function forces all probabilities in a row to sum to $1$, the presence of multiple valid positive modalities leads to competition between positives. To avoid this limitation, we adopt the sigmoid-based binary cross-entropy (BCE) loss from  SigLIP~\citep{siglip}, which treats every pair in the similarity matrix as an independent binary classification problem. We extend this SigLIP formulation to our multimodal setting with $M$ modalities.

For each sample $n$, we form an $M \times M$ block of modality--modality relationships. $\mathbf{x}_{n,i} \in \mathbb{R}^C$ denotes the feature of modality $i$ for sample $n$. For every possible modality pair $(i,j)$, we associate a dedicated projection parameterized by a weight matrix $\mathbf{W}_{i,j} \in \mathbb{R}^{C\times C}$ and a bias vector $\mathbf{b}_{i,j} \in \mathbb{R}^{C}$. The projected embedding from modality $i$ to modality $j$ is computed as
\[
\mathbf{h}_{n, i \rightarrow j}
= \texttt{Normalize} (\mathbf{W}_{i,j} \, \mathbf{x}_{n,i} + \mathbf{b}_{i,j}).
\]

Projecting each source modality into the representational space of every target modality enables the model to learn modality-pair-specific interaction spaces between biological layers. Similarities are computed across the global batch, yielding modality-pair-specific logits  
\[
\ell_{n,i,k,j} = \alpha_{i,j} \cdot \langle \mathbf{h}_{n, i \rightarrow j}, \mathbf{h}_{k, j \rightarrow i} \rangle + \beta_{i,j},
\]
where $\alpha_{i,j}$ and bias $\beta_{i,j}$ are learnable modality-dependent scaling and bias parameters. 
Each logit is then compared against a binary target $t_{n,i,k,j} \in \{0,1\}$ indicating whether modality $i$ of sample $n$ forms a valid positive pair with modality $j$ of sample $k$: 
\[
t_{n,i,k,j} = \mathbf1 (n \equiv k) \times \texttt{ValidPair}_{n,i,j}, 
\]
where $\texttt{ValidPair}_{n,i,j}$ denotes whether modalities $i$, $j$ are both available for sample $n$. Positives occur only within the same sample while negatives are collected across the global batch. The loss for each modality pair is computed as 
\[
\mathcal{L}_{n,i,k,j}
= \texttt{BCEWithLogits}
\bigl(\ell_{n,i,k,j},\, t_{n,i,k,j}\bigr).
\]

Overall, our modified Multimodal SigLIP loss allows every modality pair $(i,j)$ to contribute to an independent BCE term, with class balancing achieved by averaging positive and negative contributions.

\subsection{Fragment-Aware Rotary Position Encoding (F-RoPE)}

High resolution whole-slide images are typically processed as a set of local patches. While patch-level encoders capture local morphology, naively aggregating these features ignores the spatial organization of tissue and may blur biologically meaningful structures. At the same time, multiple separated regions on a single slide are common in routine histopathology workflows, where several tissue sections from the same patient (e.g., different biopsy site, stain type) are stained together~\cite{allen2006histopathology, khened2021generalized}. Hence, F-RoPE aims to consider the decomposition of slides into separate tissue fragments.

To encode spatial relationships between patches while avoiding spurious positional coupling across physically disconnected regions, we augment the WSI aggregator with (1) RoPE-based positional encoding~\citep{rope} and (2) a fragment-aware attention mask. For each patch, we maintain its pixel coordinate and the fragment index obtained from tissue segmentation. Pixel coordinates are converted into continuous patch indices by normalizing with respect to the patch size, and then passed through a RoPE module to obtain sinusoidal position embeddings that rotate query/key vectors.

In parallel, we construct a pairwise attention mask over all tokens. Each patch token is assigned its corresponding fragment index, whereas the CLS and register tokens are grouped together and assigned to a separate single fragment. Based on this fragment index, we define a boolean mask by the equality of these indices. Then this fragment-aware mask is used to restrict attention in the RoPE branch to preserve only interactions within the fragment and suppress positional attention across disconnected tissue fragments.

Our aggregator transformer interleaves RoPE-attention and NoPE-attention blocks.
The RoPE blocks incorporate rotary positional embeddings together with the fragment-aware attention mask to capture local spatial geometry and fragment-level topology.
In contrast, the NoPE blocks remain fully position-agnostic over valid tokens relying solely on the padding masks.
This alternating dual-attention scheme enables the model to learn both localized structural cues and global slide-level evidence without spatial bias, achieving a balanced and expressive WSI representation.

\subsection{Overall Architecture: EXAONE Path 2.5}

\paragraph{WSI Encoder.}
The WSI encoder $E_{\text{WSI}}$ consists of (1) an internally pretrained WSI patch-level foundation model, (2) a Transformer-based aggregator equipped with F-RoPE, and (3) a linear projection head.
WSIs are first encoded into patch embeddings using the WSI foundation model. These patch tokens are then processed with its patch coordinates through the aggregator module. 
The spatial fragment indices refine the final slide embedding to encompass fragment-level structural consistency.

\paragraph{Gene Encoders.}
We assign four separate gene encoders ($E_{\text{RNA}}, E_{\text{SNP}}, E_{\text{CNV}}, E_{\text{Meth}}$) to individually process each gene modality showing different degrees of sparsity and expressivity. RNA expression profiles are encoded using an internally pretrained RNA-seq foundation model, whereas SNP, CNV, DNA methylation expression values are encoded as fixed-length vector representations. Each encoder is a feed-forward module composed of an initial linear projection followed by several pre-norm SwiGLU FFN residual blocks.

\paragraph{Multimodal Embedding and Alignment.}
The final embeddings from all modalities are aligned using the Multimodal SigLIP loss (Section~\ref{sec:siglip}), which supervises all modality pairs jointly. This pairwise alignment formulation encourages biologically meaningful cross-modal consistency while preserving modality-specific representations in a shared embedding space.

\section{Experiments}

\subsection{Training Data}
Our multimodal training cohort consists of five complementary data modalities: whole-slide images, bulk RNA sequencing, single-nucleotide polymorphisms, copy-number variations, and DNA methylation profiles. We have curated a total of 23,099 patients who have at least two modalities available. The final dataset used to train our multimodal contrastive learning framework and WSI feature aggregator includes 23,099 WSI slides (one per patient), 29,876 RNA-seq samples, 10,250 SNP samples, 10,601 CNV samples, and 10,503 DNA methylation samples. The WSI and RNA foundation models are pretrained and used to extract patch-level and RNA-sequence embeddings while kept frozen throughout this process.

\subsection{Benchmarks}
To assess both real-world performance and generalizability, we evaluate our model using (1) internal clinical datasets collected from multiple hospitals and (2) the public Patho-Bench\footnote{\url{https://huggingface.co/datasets/MahmoodLab/Patho-Bench}} benchmark suite.

\paragraph{(1) Internal Clinical Benchmarks.}
We have curated internal datasets from one general hospital in Korea~(KOR) and two general hospitals in the United States (USA1, USA2). LUAD-TMB task predicts tumor mutation burden (TMB) status (high vs.\ low) defined according to the number of mutations per megabase with a threshold of 10 from lung adenocarcionma (LUAD) patients. \mbox{LUAD-EGFR} and LUAD-KRAS identifies the presence of actionable EGFR and KRAS mutations in LUAD respectively. CRC-MSI task classifies microsatellite instability (MSI) status in colorectal adenocarcinoma (CRC). All tasks are binary classification tasks. These seven internal benchmarks evaluate cross-institution generalization clinical robustness of the learned WSI representations to transfer to real-world diagnostic and molecular prediction tasks.
All data usage has been approved by the respective Institutional Review Boards (IRBs) and fully de-identified. The dataset is restricted for internal research use. 
 
\paragraph{(2) Public Benchmark: Patho-Bench.}
We further evaluate all models on Patho-Bench, a large-scale and comprehensive benchmark suite for computational pathology. Patho-Bench spans across seven major clinical and biological task families: morphological subtyping, tumor microenvironment (TME) characterization, tumor grading, molecular subtyping, mutation prediction, treatment response prediction, and survival prediction. The dataset provides task-specific labels across diverse cancer types, enabling unified and systematic evaluation of WSI representation learning methods.

The Patho-Bench tasks utilized in our evaluation setting after excluding unavailable or corrupted datasets are shown in Table~\ref{tab:pathobench_tasks} and Table~\ref{tab:pathobench_result}. For all experiments, we strictly follow the official 5-fold/50-fold cross-validation protocol provided by Patho-Bench. Each task includes predefined train/test splits, ensuring fair and consistent comparison with prior works. Performance is reported as the average across folds for each task, and all metrics are computed using the official evaluation scripts provided in THREADS~\citep{threads}.

\subsection{Baseline Models}
We compare our method against six state-of-the-art pathology and multimodal foundation models with publicly available checkpoints. For all baselines, we adopt their official preprocessing pipelines and evaluate the released encoders without any additional fine-tuning.

\paragraph{UniModal Foundation Models.}
CHIEF~\cite{chief} is a contrastive pathology foundation model pretrained on 60K whole-slide images (WSIs) using a CTransPath~\cite{ctranspath} backbone coupled with an ABMIL~\cite{abmil} aggregator. Prov-GigaPath~\cite{gigapath} is trained on 171K slides from the Prov-Path dataset using a DINOv2~\cite{dinov2} objective with a ViT-G backbone.
We additionally include two patch-level encoders that do not provide slide-level representations. H-optimus-0\footnote{\url{https://huggingface.co/bioptimus/H-optimus-0}} is a ViT-based encoder trained on over 500K WSIs and UNI~\cite{UNI} is a large-scale self-supervised model pretrained on more than 100 million WSIs.

\paragraph{Multimodal Foundation Models.}
TITAN~\cite{titan} adopts a teacher–student self-distillation framework built on the pretrained CONCHv1.5\footnote{\url{https://huggingface.co/MahmoodLab/conchv1_5}} encoder and is trained on its internal MASS-340K dataset comprising WSIs, generated region level captions and pathology reports. PRISM \cite{prism} also integrates WSIs and pathology reports using a Perceiver architecture~\cite{jaegle2021perceiver} combined with the Virchow~\cite{virchow} pathology encoder.

\subsection{Evaluation Protocols} \label{sec:eval_protocol}
\begin{table}[t]
\centering
\caption{Internal Clinical Benchmark - Train : Test Split.}
\resizebox{350px}{!}{%
\begin{tabular}{lcccc}
\toprule
\textbf{Task} & \textbf{Train (KOR)} & \textbf{Test (USA1)} & \textbf{Test (USA2)} & \textbf{Test (KOR)}\\
\midrule
\textbf{LUAD-TMB} (high : low)  & 1063 : 287 & 137 : 117 & 308 : 207 & - \\
\textbf{LUAD-EGFR} (wild : mut) & 1145 : 205 & 242 : 12  & 426 : 89 & - \\
\textbf{LUAD-KRAS} (wild : mut) & 1217 : 133 & 163 : 91  & 347 : 168 & - \\
\textbf{CRC-MSI} (stable : instable) & 2630 : 831 & - & - & 658 : 209 \\
\bottomrule
\end{tabular}}
\label{tab:internal}
\end{table}
\paragraph{Internal Clinical Benchmarks.}
For patch-level foundation models (H-optimus-0, UNI-2h), we follow UNI~\cite{UNI} and obtain slide-level representations using a CLAM aggregator~\cite{clam}.
For slide-level foundation models (CHIEF, GigaPath, PRISM, TITAN), we train a linear classifier on top of their frozen slide-level embeddings.
EXAONEPath 2.5 is compared in both evaluation settings.
We perform a learning-rate search and report the averaged AUROC across four random seed runs, selecting the best-performing learning rate for each model.
All models are trained and evaluated using the predefined dataset splits shown in Table~\ref{tab:internal}.

\begin{table}[t]
\centering
\caption{Distribution of evaluation tasks in Patho-Bench.}
\label{tab:pathobench_tasks}
\begin{tabular}{l c}
\toprule
\textbf{Task Category} & \textbf{\# Tasks} \\
\midrule
Mutation prediction & 30 \\
Tumor grading & 7 \\
Tumor microenvironment (TME) characterization & 14 \\
Treatment response and assessment & 7 \\
Molecular subtyping & 7 \\
Morphological subtyping & 6 \\
Survival analysis & 9 \\
\midrule
\textbf{Total} & \textbf{80} \\
\bottomrule
\end{tabular}
\end{table}

\paragraph{Patho-Bench Classification Tasks.}
We evaluate patch-level foundation models using mean pooling of patch embeddings followed by a linear classifier.
For slide-level foundation models, we directly train a linear classifier on their released slide-level embeddings. Linear probing follows the logistic regression configuration used in THREADS with fixed hyperparameters~\cite{threads}. All classification tasks are evaluated using macro-averaged AUROC (one-vs-rest). 

\paragraph{Patho-Bench Time-to-Event Prediction Tasks.}
We evaluate both patch-level and slide-level foundation models using a unified survival-probing framework. For slide-level models, we directly feed their released slide embeddings into a Cox proportional hazards model. For patch-level models, we first construct slide representations by mean pooling of patch embeddings and then train a Cox model on the resulting slide features. All Cox models use the same fixed hyperparameters as in THREADS, and performance is evaluated using the concordance index (C-index).

\section{Results}

\subsection{Performance on Internal Clinical Benchmarks}
\label{sec:internal_results}

\begin{figure}[t]
    \centering  \includegraphics[width=0.75\linewidth]{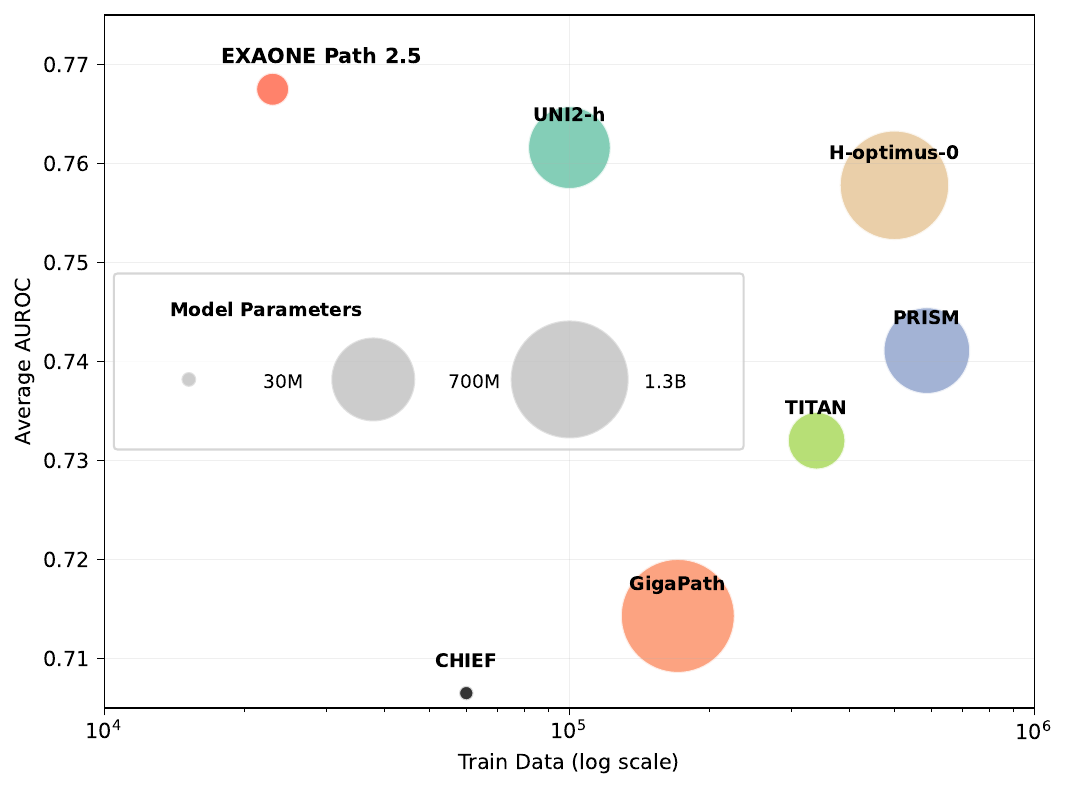}
    \caption{Internal benchmark performance of pathology foundation models. Marker area is proportional to each model’s parameter count, and the $x$-axis shows the pretraining data scale (log). EXAONE Path 2.5 (CLAM) delivers the best AUROC while being among the smallest models trained on the least data, outperforming billion-parameter models pretrained on substantially larger datasets.}
    \label{fig:internal_benchmark}
\end{figure}
We evaluate EXAONE Path 2.5 on internal multi-institutional benchmarks covering three lung adenocarcinoma tasks (TMB10, EGFR, KRAS) and one colorectal MSI task, each assessed across the Korean (KOR) cohort and two U.S. hospital cohorts (USA1, USA2). Table~\ref{tab:internal_benchmark} reports AUROC performance against six representative pathology foundation models: TITAN, PRISM, CHIEF, GigaPath, UNI2-h, and H-optimus-0.

EXAONE Path 2.5 achieves the highest average AUROC (0.7675) across all tasks, demonstrating robust generalization under both evaluation protocols. As described in Section~\ref{sec:eval_protocol}, slide-level foundation models (blue) are evaluated with a linear classifier, whereas patch-level models (red) require CLAM-based aggregation prior to classification. EXAONE Path 2.5 was assessed in both settings and consistently ranked first within each protocol.

Figure~\ref{fig:internal_benchmark} further characterizes performance with respect to pretraining data scale and model size. Despite having substantially fewer parameters and being pretrained on less data than billion-parameter baselines, EXAONE Path 2.5 achieves superior AUROC. This highlights that architectural choices tailored to pathology—such as fragment-aware positional encoding and multimodal alignment throughout multiple biological layers—can provide greater gains than scale alone, yielding efficient and effective slide representations.

\begin{table*}[t]
\centering
\scriptsize
\renewcommand{\arraystretch}{1.25}
\setlength{\tabcolsep}{4pt}

\caption{Performance comparison across tasks and datasets. Blue cells correspond to models evaluated in the linear setting, while red cells correspond to the CLAM setting.
EXAONE Path 2.5 is evaluated in both settings, and each of its configurations (linear and CLAM) achieves the best performance within its respective evaluation regime.
Row-wise best values are shown in bold.}

\begin{tabular}{
lll | 
>{\columncolor{blue!10}}c
>{\columncolor{blue!10}}c
>{\columncolor{blue!10}}c
>{\columncolor{blue!10}}c | 
>{\columncolor{purple!10}}c
>{\columncolor{purple!10}}c | 
>{\columncolor{blue!10}}c
>{\columncolor{purple!10}}c
}

\toprule
\textbf{Cancer} & \textbf{Target} & \textbf{Test set} &
\cellcolor{white}\textbf{TITAN} & \cellcolor{white}\textbf{PRISM} & 
\cellcolor{white}\textbf{CHIEF} & \cellcolor{white}\textbf{GigaPath} &
\cellcolor{white}\textbf{UNI2-h} & \cellcolor{white}\textbf{H-optimus-0} & 
\multicolumn{2}{c}{\textbf{EXAONE Path 2.5}} \\
\midrule

\multirow{6}{*}{LUAD}
& \multirow{2}{*}{TMB10}
& USA1   & 0.6897 & 0.6664 & 0.6505 & 0.6765 & 0.6893 & 0.6590 & \textbf{0.6936} & 0.6824 \\
& & USA2& 0.6904 & 0.6857 & 0.6546 & 0.6731  & 0.6860 & 0.6928 & \textbf{0.7077} & 0.6951 \\

& \multirow{2}{*}{EGFR}
& USA1    & 0.7839 & 0.8296 & 0.8109 & 0.7073  & 0.8192 & 0.8508 & 0.7843 & \textbf{0.8555} \\
& & USA2 & 0.8278 & 0.8163 & 0.7690 & 0.7629  & \textbf{0.8628} & 0.8535 & 0.8274 & 0.8482 \\

& \multirow{2}{*}{KRAS}
& USA1    & 0.5895 & 0.5749 & 0.5422 & \textbf{0.6167}  & 0.6075 & 0.5995 & 0.5534 & 0.6022 \\
& & USA2 & 0.6104 & 0.6796 & 0.5987 & 0.6149 & 0.6856 & 0.6382 & 0.6983 & \textbf{0.7102} \\
\midrule

CRC
& MSI
& KOR & 0.9320 & 0.9352 & 0.9199 & 0.9499 &\textbf{ 0.9807} & 0.9907 & 0.9609 & 0.9791 \\
\midrule

\multicolumn{3}{c|}{\textbf{AVERAGE}} 
& 0.7320 & 0.7411 & 0.7065 & 0.7143 
& 0.7616 & 0.7578 & 0.7465 & \textbf{0.7675} \\
\bottomrule

\end{tabular}

\label{tab:internal_benchmark}
\end{table*}

\subsection{Performance on Patho-Bench}

\begin{figure}[t]
    \centering  \includegraphics[width=0.75\linewidth]{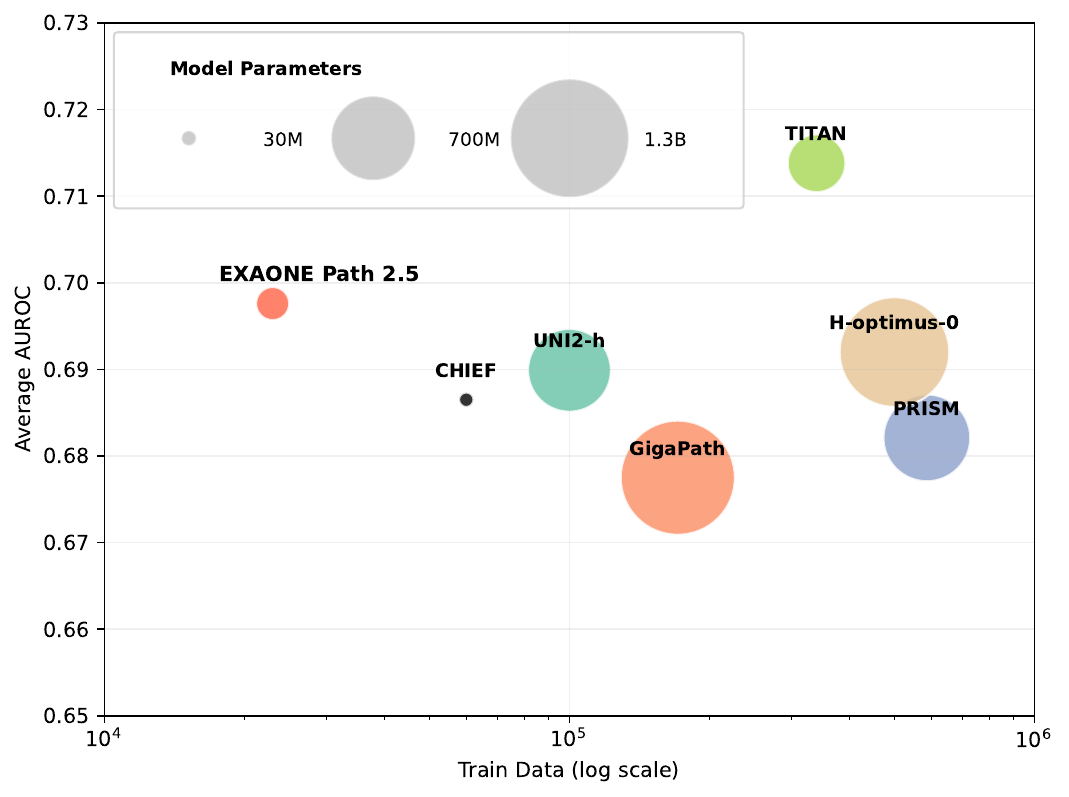}
    \caption{Performance of pathology foundation models on Patho-Bench. The plot shows average AUROC on Patho-Bench consisting of 80 tasks (5/50 fold evaluation). While prior state-of-the-art models benefit from large-scale training corpora, EXAONE Path 2.5 delivers comparable performance under a substantially smaller training regime, demonstrating favorable data–performance trade-offs.}
    \label{fig:pathobench}
\end{figure}

To measure the comprehensive capability of foundation models and ensure reproducible results, EXAONE Path 2.5 is evaluated using Patho-Bench evaluation protocol. The Patho-Bench evaluation includes 26 public datasets composed of 24,483 WSIs and covers 80 tasks. Experiments are conducted with consistent settings across all baseline models. For patch-level foundation models, we apply mean pooling to aggregate features for fair comparison with slide-level approaches. 

Despite substantial differences in training data scale and model size, EXAONE Path 2.5 performs on par with state-of-the-art models on Patho-Bench. EXAONE Path 2.5 demonstrates not only superior performance on out-of-distribution data but also stable and comprehensive performance across in-distribution evaluations.

\begin{figure}[t]
    \centering  \includegraphics[width=\linewidth]{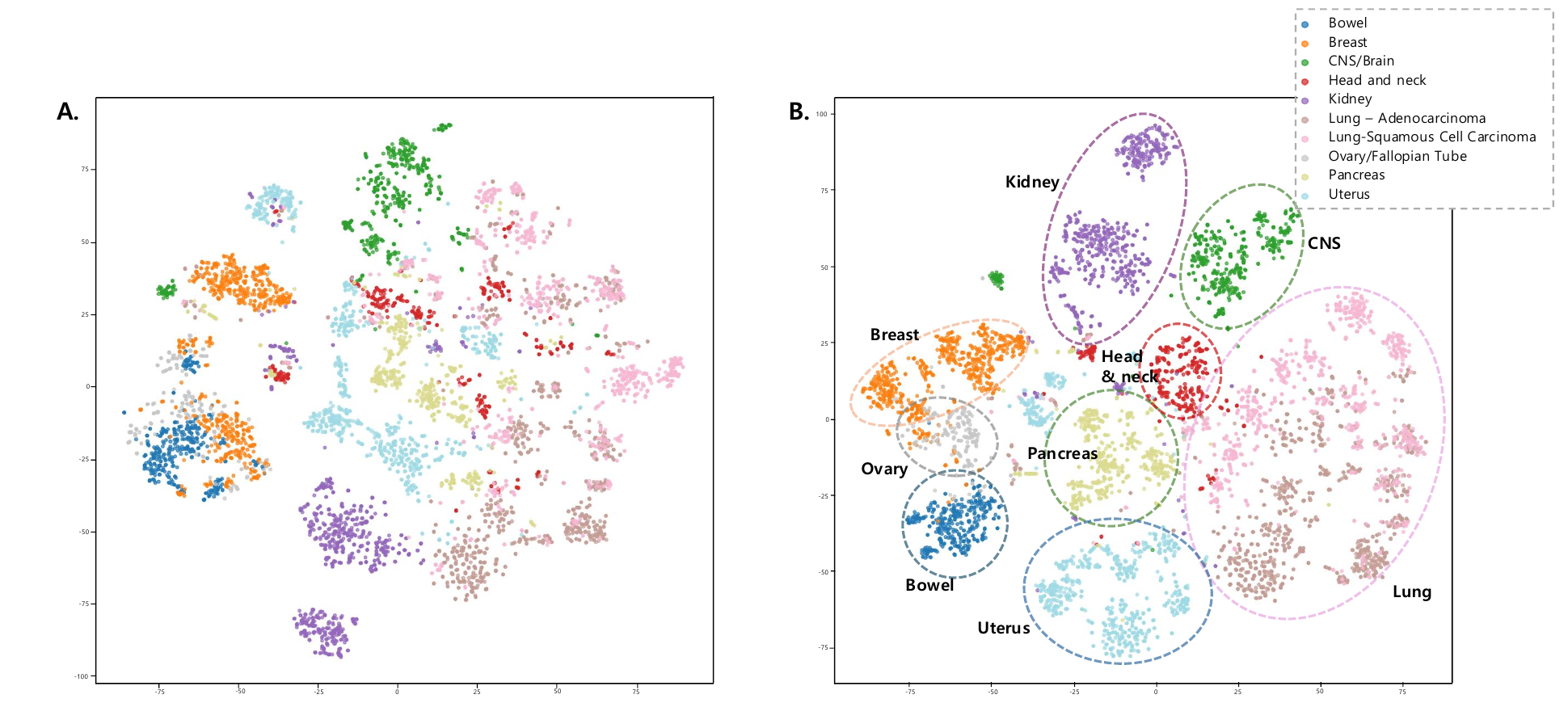}
    \caption{(A) t-SNE embedding of slide-level WSI patch features derived from the internal pathology foundation model.
    (B) t-SNE embedding of the same samples after cross-modal alignment with molecular features (RNA-seq, SNP, CNV, DNA methylation) using EXAONE Path 2.5.
    Integrating molecular information yields substantially improved cluster organization and enhanced separation across major anatomical and histological categories.}
    \vspace{-5px}
    \label{fig:visualization}
\end{figure}

To investigate how multi-omics alignment enhances the biological expressiveness of histopathology representations, we analyze the intrinsic structure of learned patch embeddings through unsupervised clustering. Specifically, (1) patch embeddings directly extracted from the internally pretrained WSI foundation model and (2) patch embeddings obtained after multimodal alignment with molecular modalities are compared. All whole-slide images are embedded from ten CPTAC cohorts spanning nine distinct organ types. Patch-level embeddings are aggregated into slide-level representations using simple mean pooling, and the resulting latent space is visualized using t-SNE.
As shown in Figure~\ref{fig:visualization}, embeddings learned with multimodal alignment exhibit markedly improved organ-wise clustering compared to unimodal baselines, indicating that incorporating molecular context amplifies biologically meaningful signals in histopathologic features and yields representations that better reflect underlying tissue and organ-specific phenotypes.

\section{Conclusion}

In this work, we introduced EXAONE Path 2.5, a pathology foundation model that bridges histopathology and multi-omics data to construct a biologically grounded representation of cancer from genotype to phenotype. By jointly modeling whole-slide images together with genomic, epigenetic, and transcriptomic modalities, our framework moves beyond morphology-only learning and captures complementary biological signals that are critical for understanding tumor behavior.

To achieve robust and efficient multimodal integration, we proposed three key design components: (1) a multimodal SigLip loss that enables all-pairwise contrastive alignment across heterogeneous modalities without inducing competition among positives, (2) a fragment-aware rotary positional encoding (F-RoPE) mechanism that preserves spatial structure and tissue-fragment topology in whole-slide images, and (3) the use of domain-specialized internal foundation models for WSI and RNA-seq to provide stable and biologically meaningful embeddings for multimodal alignment. Together, these components allow EXAONE Path 2.5 to effectively integrate diverse biological layers while remaining data- and parameter-efficient.

Extensive experiments on both internal multi-institutional clinical benchmarks and the Patho-Bench benchmark comprising 80 tasks demonstrate that EXAONE Path 2.5 achieves performance superior or on par with state-of-the-art pathology foundation models, despite being trained with substantially fewer parameters and less pretraining data. Notably, the model exhibits strong generalization across institutions and cancer types, as well as improved biological coherence in the learned representation space, as evidenced by enhanced organ- and tissue-level clustering after multimodal alignment.

Overall, our results highlight that biologically informed multimodal design can provide greater gains than scale alone in computational pathology. EXAONE Path 2.5 underscores the potential of integrated genotype-to-phenotype modeling as a foundation for next-generation precision oncology, enabling more robust molecular inference, improved clinical generalization, and deeper biological interpretability from routine histopathology data.

\clearpage
\begin{table*}[t]
\centering

\caption{Performance comparison across Patho-Bench tasks. The task column follows official Patho-Bench task names, and all models are evaluated under the same settings using linear probing. H-optimus-0 and UNI2-h use mean pooling to obtain slide-level features.}

\resizebox{\textwidth}{!}{%
\begin{tabular}{
l | c c c c c c c c
}

\toprule
\textbf{Task} &
\textbf{CHIEF} & \textbf{GigaPath} & 
\textbf{PRISM} & \textbf{TITAN} &
\textbf{H-optimus-0} & \textbf{UNI2-h} & 
\textbf{EXAONE Path 2.5} \\
\midrule

bc\_therapy\_er\_status & 0.664 & 0.697 & 0.625 & 0.763 & 0.683 & 0.711 & 0.658 \\
bc\_therapy\_grade & 0.743 & 0.654 & 0.722 & 0.781 & 0.731 & 0.751 & 0.720 \\
bc\_therapy\_her2\_status & 0.644 & 0.646 & 0.718 & 0.729 & 0.711 & 0.678 & 0.735 \\
bc\_therapy\_residual\_cancer\_burden & 0.571 & 0.553 & 0.572 & 0.592 & 0.577 & 0.590 & 0.626 \\
bcnb\_er & 0.872 & 0.895 & 0.893 & 0.919 & 0.928 & 0.912 & 0.914 \\
bcnb\_her2 & 0.722 & 0.719 & 0.718 & 0.749 & 0.747 & 0.747 & 0.708 \\
bcnb\_pr & 0.815 & 0.814 & 0.821 & 0.836 & 0.858 & 0.851 & 0.847 \\
boehmk\_\_PFS & 0.454 & 0.527 & 0.541 & 0.530 & 0.478 & 0.549 & 0.500 \\
bracs\_slidelevel\_coarse & 0.884 & 0.817 & 0.893 & 0.911 & 0.831 & 0.836 & 0.832 \\
bracs\_slidelevel\_fine & 0.834 & 0.765 & 0.839 & 0.850 & 0.786 & 0.793 & 0.793 \\
cptac\_brca\_Immune\_class & 0.680 & 0.708 & 0.643 & 0.687 & 0.716 & 0.694 & 0.707 \\
cptac\_brca\_PIK3CA\_mutation & 0.623 & 0.497 & 0.612 & 0.585 & 0.566 & 0.523 & 0.545 \\
cptac\_brca\_TP53\_mutation & 0.820 & 0.740 & 0.853 & 0.838 & 0.748 & 0.766 & 0.762 \\
cptac\_ccrcc\_BAP1\_mutation & 0.641 & 0.662 & 0.587 & 0.677 & 0.618 & 0.697 & 0.612 \\
cptac\_ccrcc\_Immune\_class & 0.565 & 0.596 & 0.558 & 0.605 & 0.572 & 0.582 & 0.590 \\
cptac\_ccrcc\_OS & 0.435 & 0.507 & 0.571 & 0.482 & 0.614 & 0.378 & 0.554 \\
cptac\_ccrcc\_PBRM1\_mutation & 0.470 & 0.481 & 0.504 & 0.528 & 0.447 & 0.473 & 0.454 \\
cptac\_ccrcc\_VHL\_mutation & 0.561 & 0.602 & 0.462 & 0.522 & 0.486 & 0.459 & 0.524 \\
cptac\_coad\_ACVR2A\_mutation & 0.762 & 0.702 & 0.725 & 0.802 & 0.779 & 0.798 & 0.849 \\
cptac\_coad\_APC\_mutation & 0.613 & 0.713 & 0.694 & 0.748 & 0.702 & 0.722 & 0.781 \\
cptac\_coad\_ARID1A\_mutation & 0.777 & 0.783 & 0.764 & 0.754 & 0.792 & 0.745 & 0.806 \\
cptac\_coad\_Immune\_class & 0.590 & 0.631 & 0.629 & 0.592 & 0.593 & 0.613 & 0.630 \\
cptac\_coad\_KRAS\_mutation & 0.687 & 0.671 & 0.609 & 0.607 & 0.622 & 0.679 & 0.652 \\
cptac\_coad\_MSI\_H & 0.844 & 0.849 & 0.859 & 0.875 & 0.885 & 0.884 & 0.933 \\
cptac\_coad\_PIK3CA\_mutation & 0.643 & 0.705 & 0.510 & 0.647 & 0.688 & 0.668 & 0.675 \\
cptac\_coad\_SETD1B\_mutation & 0.711 & 0.707 & 0.732 & 0.798 & 0.790 & 0.808 & 0.901 \\
cptac\_coad\_TP53\_mutation & 0.664 & 0.644 & 0.675 & 0.714 & 0.718 & 0.734 & 0.770 \\
cptac\_gbm\_EGFR\_mutation & 0.734 & 0.629 & 0.659 & 0.747 & 0.562 & 0.619 & 0.613 \\
cptac\_gbm\_Immune\_class & 0.644 & 0.607 & 0.706 & 0.672 & 0.646 & 0.656 & 0.633 \\
cptac\_gbm\_TP53\_mutation & 0.825 & 0.720 & 0.811 & 0.800 & 0.749 & 0.752 & 0.762 \\
cptac\_hnsc\_CASP8\_mutation & 0.409 & 0.424 & 0.630 & 0.523 & 0.394 & 0.366 & 0.516 \\
cptac\_hnsc\_Histologic\_Grade & 0.764 & 0.740 & 0.712 & 0.758 & 0.777 & 0.797 & 0.785 \\
cptac\_hnsc\_Immune\_class & 0.626 & 0.595 & 0.712 & 0.612 & 0.620 & 0.614 & 0.623 \\
cptac\_hnsc\_OS & 0.565 & 0.664 & 0.599 & 0.578 & 0.678 & 0.599 & 0.738 \\
cptac\_lscc\_ARID1A\_mutation & 0.478 & 0.433 & 0.432 & 0.491 & 0.320 & 0.365 & 0.349 \\
cptac\_lscc\_Histologic\_Grade & 0.641 & 0.705 & 0.602 & 0.621 & 0.705 & 0.708 & 0.695 \\
cptac\_lscc\_Immune\_class & 0.675 & 0.686 & 0.643 & 0.755 & 0.686 & 0.671 & 0.655 \\
cptac\_lscc\_KEAP1\_mutation & 0.627 & 0.576 & 0.531 & 0.553 & 0.646 & 0.632 & 0.624 \\
cptac\_luad\_EGFR\_mutation & 0.710 & 0.813 & 0.802 & 0.786 & 0.803 & 0.782 & 0.780 \\
cptac\_luad\_Immune\_class & 0.720 & 0.632 & 0.660 & 0.697 & 0.625 & 0.613 & 0.644 \\
cptac\_luad\_KRAS\_mutation & 0.630 & 0.743 & 0.682 & 0.742 & 0.759 & 0.699 & 0.723 \\
cptac\_luad\_OS & 0.466 & 0.471 & 0.462 & 0.666 & 0.581 & 0.500 & 0.481 \\
cptac\_luad\_STK11\_mutation & 0.843 & 0.828 & 0.854 & 0.869 & 0.859 & 0.859 & 0.839 \\
cptac\_luad\_TP53\_mutation & 0.672 & 0.687 & 0.718 & 0.740 & 0.678 & 0.731 & 0.634 \\
cptac\_ov\_Immune\_class & 0.597 & 0.521 & 0.558 & 0.572 & 0.522 & 0.509 & 0.553 \\
cptac\_pda\_Immune\_class & 0.495 & 0.537 & 0.544 & 0.533 & 0.538 & 0.534 & 0.536 \\
cptac\_pda\_OS & 0.546 & 0.454 & 0.590 & 0.591 & 0.490 & 0.457 & 0.505 \\
cptac\_pda\_SMAD4\_mutation & 0.417 & 0.467 & 0.568 & 0.537 & 0.484 & 0.416 & 0.491 \\
cptac\_ucec\_CTNNB1\_mutation & 0.714 & 0.748 & 0.718 & 0.713 & 0.756 & 0.748 & 0.767 \\
cptac\_ucec\_Immune\_class & 0.569 & 0.572 & 0.654 & 0.679 & 0.582 & 0.607 & 0.587 \\
cptac\_ucec\_PTEN\_mutation & 0.574 & 0.620 & 0.691 & 0.633 & 0.666 & 0.686 & 0.735 \\
ebrains\_diagnosis & 0.975 & 0.978 & 0.984 & 0.986 & 0.985 & 0.985 & 0.981 \\
ebrains\_diagnosis\_group & 0.985 & 0.992 & 0.990 & 0.992 & 0.995 & 0.994 & 0.994 \\
ebrains\_idh\_status & 0.960 & 0.981 & 0.987 & 0.982 & 0.983 & 0.983 & 0.984 \\
hancook\_grading\_SCC\_Conventional\_Keratinizing & 0.725 & 0.683 & 0.651 & 0.722 & 0.702 & 0.722 & 0.717 \\
hancook\_grading\_SCC\_Conventional\_NonKeratinizing & 0.679 & 0.642 & 0.437 & 0.710 & 0.721 & 0.727 & 0.722 \\
hancook\_lymphovascular\_invasion\_L & 0.647 & 0.644 & 0.602 & 0.699 & 0.678 & 0.673 & 0.659 \\
hancook\_OS\_treatment\_rdc & 0.500 & 0.592 & 0.646 & 0.590 & 0.612 & 0.500 & 0.584 \\
hancook\_perineural\_invasion\_Pn & 0.731 & 0.706 & 0.691 & 0.731 & 0.724 & 0.715 & 0.724 \\
hancook\_primarily\_metastasis & 0.682 & 0.632 & 0.651 & 0.766 & 0.678 & 0.700 & 0.662 \\
hancook\_primary\_tumor\_site & 0.861 & 0.917 & 0.891 & 0.922 & 0.931 & 0.932 & 0.918 \\
hancook\_vascular\_invasion\_V & 0.645 & 0.636 & 0.627 & 0.665 & 0.666 & 0.685 & 0.678 \\
herroi\_response & 0.683 & 0.631 & 0.574 & 0.668 & 0.660 & 0.640 & 0.630 \\
imp\_grade & 0.988 & 0.987 & 0.992 & 0.995 & 0.986 & 0.987 & 0.984 \\
mbc\_\_OS & 0.575 & 0.508 & 0.508 & 0.631 & 0.495 & 0.443 & 0.507 \\
mbc\_\_Recist & 0.635 & 0.588 & 0.616 & 0.642 & 0.598 & 0.604 & 0.630 \\
mut-het-rcc\_BAP1\_mutation & 0.852 & 0.842 & 0.851 & 0.861 & 0.885 & 0.891 & 0.874 \\
mut-het-rcc\_PBRM1\_mutation & 0.781 & 0.792 & 0.803 & 0.797 & 0.820 & 0.830 & 0.815 \\
mut-het-rcc\_SETD2\_mutation & 0.738 & 0.713 & 0.714 & 0.727 & 0.730 & 0.742 & 0.710 \\
nadt\_response & 0.639 & 0.641 & 0.597 & 0.623 & 0.684 & 0.651 & 0.689 \\
natbrca\_lymphovascular\_invasion & 0.566 & 0.572 & 0.470 & 0.662 & 0.590 & 0.562 & 0.551 \\
ovarian\_response & 0.708 & 0.689 & 0.614 & 0.596 & 0.685 & 0.679 & 0.512 \\
panda\_isup\_grade & 0.944 & 0.932 & 0.949 & 0.932 & 0.949 & 0.946 & 0.956 \\
sr386\_\_braf\_mutant\_binary & 0.778 & 0.699 & 0.641 & 0.741 & 0.671 & 0.712 & 0.753 \\
sr386\_\_died\_within\_5\_years & 0.703 & 0.605 & 0.648 & 0.718 & 0.666 & 0.690 & 0.650 \\
sr386\_\_mmr\_loss\_binary & 0.844 & 0.805 & 0.836 & 0.883 & 0.841 & 0.864 & 0.852 \\
sr386\_\_OS & 0.606 & 0.602 & 0.624 & 0.631 & 0.597 & 0.625 & 0.604 \\
sr386\_\_ras\_mutant\_binary & 0.696 & 0.605 & 0.625 & 0.691 & 0.627 & 0.678 & 0.683 \\
visiomel\_relapse\_noprev\_melanoma & 0.758 & 0.691 & 0.690 & 0.745 & 0.673 & 0.722 & 0.730 \\
visiomel\_relapse\_overall & 0.806 & 0.739 & 0.788 & 0.805 & 0.734 & 0.748 & 0.711 \\

\midrule

{\textbf{AVERAGE}} 
& 0.687 & 0.678 & 0.682 & 0.714 
& 0.692 & 0.690 & 0.698 \\
\bottomrule

\end{tabular}}

\label{tab:pathobench_result}
\end{table*}

\clearpage
{\fontsize{10}{10}\selectfont
\bibliography{8.reference}
}

\clearpage
\section*{EXAONEPath AI Model License Agreement 1.0 - NC}
This License Agreement (“Agreement”) is entered into between you (“Licensee”) and LG Management Development Institute Co., Ltd. (“Licensor”), governing the use of the EXAONEPath AI Model (“Model”). By downloading, installing, copying, or using the Model, you agree to comply with and be bound by the terms of this Agreement. If you do not agree to all the terms, you must not download, install, copy, or use the Model. This Agreement constitutes a binding legal agreement between the Licensee and Licensor.
\section*{1. Definitions}
\subsection*{1.1 Model}The artificial intelligence model provided by Licensor, which includes any software, algorithms, machine learning models, or related components supplied by Licensor. This definition extends to encompass all updates, enhancements, improvements, bug fixes, patches, or other modifications that may be provided by Licensor from time to time, whether automatically or manually implemented.
\subsection*{1.2 Derivatives}Any modifications, alterations, enhancements, improvements, adaptations, or derivative works of the Model created by Licensee or any third party. This includes changes made to the Model's architecture, parameters, data processing methods, or any other aspect of the Model that results in a modification of its functionality or output.
\subsection*{1.3 Output} Any data, results, content, predictions, analyses, insights, or other materials generated by the Model or Derivatives, regardless of whether they are in their original form or have been further processed or modified by the Licensee. This includes, but is not limited to, textual or numerical produced directly or indirectly through the use of the Model.
\subsection*{1.4 Licensor} LG Management Development Institute Co., Ltd., the owner, developer, and provider of the EXAONEPath AI Model. The Licensor holds all rights, title, and interest in the Model and is responsible for granting licenses to use the Model under the terms specified in this Agreement.
\subsection*{1.5 Licensee} The individual, organization, corporation, academic institution, government agency, or other entity using or intending to use the Model under the terms and conditions of this Agreement. The Licensee is responsible for ensuring compliance with the Agreement by all authorized users who access or utilize the Model on behalf of the Licensee.
\section*{2. License Grant}
\subsection*{2.1 Grant of License} Subject to the terms and conditions outlined in this Agreement, the Licensor hereby grants the Licensee a limited, non-exclusive, non-transferable, worldwide, and revocable license to:
\begin{enumerate}[label=\alph*.]
    \item Access, download, install, and use the Model solely for research purposes. This includes evaluation, testing, academic research and experimentation.
    \item Publicly disclose research results and findings derived from the use of the Model or Derivatives, including publishing papers or presentations.
    \item Modify the Model and create Derivatives based on the Model, provided that such modifications and Derivatives are used exclusively for research purposes. The Licensee may conduct experiments, perform analyses, and apply custom modifications to the Model to explore its capabilities and performance under various scenarios. If the Model is modified, the modified Model must include "EXAONEPath" at the beginning of its name.
    \item Distribute the Model and Derivatives in each case with a copy of this Agreement.
\end{enumerate}
\subsection*{2.2 Scope of License} The license granted herein does not authorize the Licensee to use the Model for any purpose not explicitly permitted under this Agreement. Any use beyond the scope of this license, including any commercial application or external distribution, is strictly prohibited unless explicitly agreed upon in writing by the Licensor.
\section*{3. Restrictions}
\subsection*{3.1 Commercial Use} The Licensee is expressly prohibited from using the Model, Derivatives, or Output for any commercial purposes, including but not limited to, developing or deploying products, services, or applications that generate revenue, whether directly or indirectly. Any commercial exploitation of the Model or its derivatives requires a separate commercial license agreement with the Licensor. Furthermore, the Licensee shall not use the Model, Derivatives or Output to develop or improve other models, except for research purposes, which is explicitly permitted.
\subsection*{3.2 Reverse Engineering} The Licensee shall not decompile, disassemble, reverse engineer, or attempt to derive the source code, underlying ideas, algorithms, or structure of the Model, except to the extent that such activities are expressly permitted by applicable law. Any attempt to bypass or circumvent technological protection measures applied to the Model is strictly prohibited.
\subsection*{3.3 Unlawful Use} The Licensee shall not use the Model and Derivatives for any illegal, fraudulent, or unauthorized activities, nor for any purpose that violates applicable laws or regulations. This includes but is not limited to the creation, distribution, or dissemination of malicious, deceptive, or unlawful content.
\subsection*{3.4 Ethical Use} The Licensee shall ensure that the Model or Derivatives is used in an ethical and responsible manner, adhering to the following guidelines:
\begin{enumerate}[label=\alph*.]
    \item The Model and Derivatives shall not be used to generate, propagate, or amplify false, misleading, or harmful information, including fake news, misinformation, or disinformation.
    \item The Model and Derivatives shall not be employed to create, distribute, or promote content that is discriminatory, harassing, defamatory, abusive, or otherwise offensive to individuals or groups based on race, gender, sexual orientation, religion, nationality, or other protected characteristics.
    \item The Model and Derivatives shall not infringe on the rights of others, including intellectual property rights, privacy rights, or any other rights recognized by law. The Licensee shall obtain all necessary permissions and consents before using the Model and Derivatives in a manner that may impact the rights of third parties.
    \item The Model and Derivatives shall not be used in a way that causes harm, whether physical, mental, emotional, or financial, to individuals, organizations, or communities. The Licensee shall take all reasonable measures to prevent misuse or abuse of the Model and Derivatives that could result in harm or injury.
\end{enumerate}

\section*{4. Ownership}
\subsection*{4.1 Intellectual Property} All rights, title, and interest in and to the Model, including any modifications, Derivatives, and associated documentation, are and shall remain the exclusive property of the Licensor. The Licensee acknowledges that this Agreement does not transfer any ownership rights to the Licensee. All trademarks, service marks, and logos associated with the Model are the property of the Licensor.
\subsection*{4.2 Output} All output generated by the Model from Licensee Data ("Output") shall be the sole property of the Licensee. Licensor hereby waives any claim of ownership or intellectual property rights to the Output. Licensee is solely responsible for the legality, accuracy, quality, integrity, and use of the Output.
\subsection*{4.3 Attribution} In any publication or presentation of results obtained using the Model, the Licensee shall provide appropriate attribution to the Licensor, citing the Model's name and version, along with any relevant documentation or references specified by the Licensor.
\section*{5. No Warranty}
\subsection*{5.1 “As-Is” Basis} The Model, Derivatives, and Output are provided on an “as-is” and “as-available” basis, without any warranties or representations of any kind, whether express, implied, or statutory. The Licensor disclaims all warranties, including but not limited to, implied warranties of merchantability, fitness for a particular purpose, accuracy, reliability, non-infringement, or any warranty arising from the course of dealing or usage of trade.
\subsection*{5.2 Performance and Reliability} The Licensor does not warrant or guarantee that the Model, Derivatives or Output will meet the Licensee’s requirements, that the operation of the Model, Derivatives or Output will be uninterrupted or error-free, or that defects in the Model will be corrected. The Licensee acknowledges that the use of the Model, Derivatives or Output is at its own risk and that the Model, Derivatives or Output may contain bugs, errors, or other limitations.
\subsection*{5.3 No Endorsement} The Licensor does not endorse, approve, or certify any results, conclusions, or recommendations derived from the use of the Model. The Licensee is solely responsible for evaluating the accuracy, reliability, and suitability of the Model for its intended purposes.
\section*{6. Limitation of Liability}
\subsection*{6.1 No Liability for Damages} To the fullest extent permitted by applicable law, in no event shall the Licensor be liable for any special, incidental, indirect, consequential, exemplary, or punitive damages, including but not limited to, damages for loss of business profits, business interruption, loss of business information, loss of data, or any other pecuniary or non-pecuniary loss arising out of or in connection with the use or inability to use the Model, Derivatives or any Output, even if the Licensor has been advised of the possibility of such damages.
\subsection*{6.2 Indemnification} The Licensee agrees to indemnify, defend, and hold harmless the Licensor, its affiliates, officers, directors, employees, and agents from and against any claims, liabilities, damages, losses, costs, or expenses (including reasonable attorneys' fees) arising out of or related to the Licensee's use of the Model, any Derivatives, or any Output, including any violation of this Agreement or applicable laws. This includes, but is not limited to, ensuring compliance with copyright laws, privacy regulations, defamation laws, and any other applicable legal or regulatory requirements.
\section*{7. Termination}
\subsection*{7.1 Termination by Licensor} The Licensor reserves the right to terminate this Agreement and revoke the Licensee’s rights to use the Model at any time, with or without cause, and without prior notice if the Licensee breaches any of the terms or conditions of this Agreement. Termination shall be effective immediately upon notice.
\subsection*{7.2 Effect of Termination} Upon termination of this Agreement, the Licensee must immediately cease all use of the Model, Derivatives, and Output and destroy all copies of the Model, Derivatives, and Output in its possession or control, including any backup or archival copies. The Licensee shall certify in writing to the Licensor that such destruction has been completed.
\subsection*{7.3 Survival} The provisions of this Agreement that by their nature should survive termination, including but not limited to, Sections 4 (Ownership), 5 (No Warranty), 6 (Limitation of Liability), and this Section 7 (Termination), shall continue to apply after termination.
\section*{8. Governing Law}
\subsection*{8.1 Governing Law} This Agreement shall be governed by and construed in accordance with the laws of the Republic of Korea, without regard to its conflict of laws principles.
\subsection*{8.2 Arbitration} Any disputes, controversies, or claims arising out of or relating to this Agreement, including its existence, validity, interpretation, performance, breach, or termination, shall be referred to and finally resolved by arbitration administered by the Korean Commercial Arbitration Board (KCAB) in accordance with the International Arbitration Rules of the Korean Commercial Arbitration Board in force at the time of the commencement of the arbitration. The seat of arbitration shall be Seoul, Republic of Korea. The tribunal shall consist of one arbitrator. The language of the arbitration shall be English.
\section*{9. Alterations}
\subsection*{9.1 Modifications} The Licensor reserves the right to modify or amend this Agreement at any time, in its sole discretion. Any modifications will be effective upon posting the updated Agreement on the Licensor’s website or through other means of communication. The Licensee is responsible for reviewing the Agreement periodically for changes. Continued use of the Model after any modifications have been made constitutes acceptance of the revised Agreement.
\subsection*{9.2 Entire Agreement} This Agreement constitutes the entire agreement between the Licensee and Licensor concerning the subject matter hereof and supersedes all prior or contemporaneous oral or written agreements, representations, or understandings. Any terms or conditions of any purchase order or other document submitted by the Licensee in connection with the Model that are in addition to, different from, or inconsistent with the terms and conditions of this Agreement are not binding on the Licensor and are void.\\

By downloading, installing, or using the EXAONEPath AI Model, the Licensee acknowledges that it has read, understood, and agrees to be bound by the terms and conditions of this Agreement.

\end{document}